\documentclass[10pt,twocolumn,notitlepage]{article} 
\usepackage{simpleConference}
\usepackage{times}
\usepackage{graphicx}
\usepackage{amssymb}
\usepackage{url,hyperref}
\usepackage{caption}
\usepackage{courier}  

\usepackage{algorithm}
\usepackage{algorithmic}
\usepackage{newfloat}
\usepackage{listings}
\floatstyle{ruled}
\newfloat{listing}{tb}{lst}{}
\floatname{listing}{Listing}

\usepackage{mathtools}
\usepackage{booktabs}    
\usepackage{multirow}
\usepackage{amsmath,amsfonts,amssymb}
\usepackage{newtxmath,newtxtext}
\usepackage[toc,page]{appendix}
\usepackage{cite}

\begin{document}

\title{DreamSalon: A Staged Diffusion Framework for Preserving Identity-Context in Editable Face Generation}

\author{
    Haonan Lin\textsuperscript{\rm 1\thanks{This work was completed during the internship at SGIT AI Lab, State Grid Corporation of China.}},
    Mengmeng Wang\textsuperscript{\rm 4,2\thanks{Corresponding author.}},
    Yan Chen\textsuperscript{\rm 1},
    Wenbin An\textsuperscript{\rm 1},
    Yuzhe Yao\textsuperscript{\rm 1}, \\
    Guang Dai\textsuperscript{\rm 2},
    Qianying Wang\textsuperscript{\rm 3},
    Yong Liu\textsuperscript{\rm 4},
    Jingdong Wang\textsuperscript{\rm 5}
}

\affiliations{
    \textsuperscript{\rm 1} MOEKLINNS Lab, Xi'an Jiaotong University, 
    \textsuperscript{\rm 2}SGIT AI Lab, State Grid Corporration of China, \\
    \textsuperscript{\rm 3}Lenovo Research,
    \textsuperscript{\rm 4}Zhejiang Univeristy, 
    \textsuperscript{\rm 5}Baidu Inc. \\
    {\tt\small \{linhaonan,wenbinan\}@stu.xjtu.edu.cn, mengmengwang@zju.edu.cn, chenyan@mail.xjtu.edu.cn,} \\
    {\tt\small \{yuzheyao.22,guang.gdai\}@gmail.com, wangqya@lenovo.com, wangjingdong@outlook.com}
%
}

\maketitle

\begin{abstract}
While large-scale pre-trained text-to-image models can synthesize diverse and high-quality human-centered images, novel challenges arise with a nuanced task of ``identity fine editing" -- precisely modifying specific features of a subject while maintaining its inherent identity and context. 
Existing personalization methods either require time-consuming optimization or learning additional encoders, adept in ``identity re-contextualization". However, they often struggle with detailed and sensitive tasks like human face editing.
To address these challenges, we introduce \textbf{DreamSalon}, a noise-guided, staged-editing framework, uniquely focusing on detailed image manipulations and identity-context preservation. 
By discerning editing and boosting stages via the frequency and gradient of predicted noises, DreamSalon first performs detailed manipulations on specific features in the editing stage, guided by high-frequency information, and then employs stochastic denoising in the boosting stage to improve image quality. For more precise editing, DreamSalon semantically mixes source and target textual prompts, guided by differences in their embedding covariances, to direct the model's focus on specific manipulation areas.
Our experiments demonstrate DreamSalon's ability to efficiently and faithfully edit fine details on human faces, outperforming existing methods both qualitatively and quantitatively.
\end{abstract}
\section{Introduction}
\label{sec:intro}

\begin{figure}[h]
    \centering
    \includegraphics[width=1.0\linewidth]{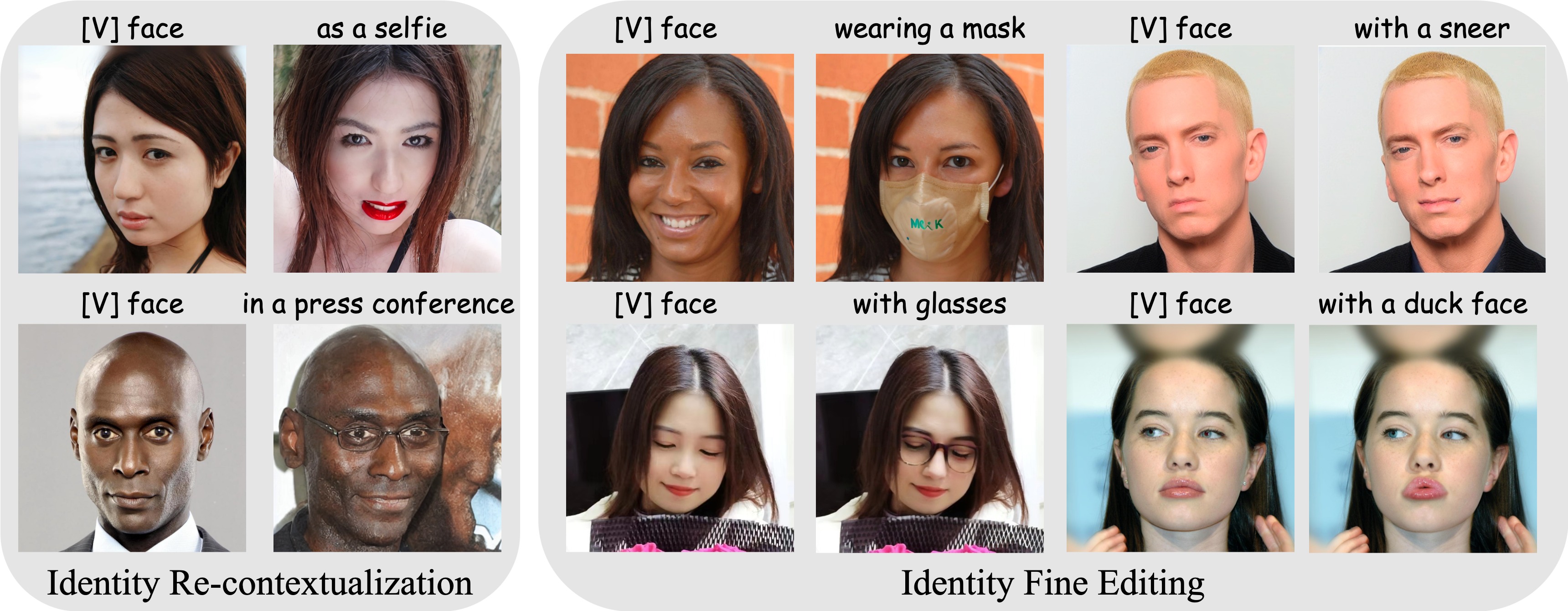}
    \caption{Unlike ``\textit{identity re-contextualization}" (Dreambooth \cite{ruiz2022dreambooth}), ``\textit{identity fine editing}" precisely manipulates details while preserving identity and context (DreamSalon).}
    \label{fig:id-fine-edit}
\end{figure}

Recent work on text-to-image (T2I) generation models has brought significant capabilities in the creation of visual content using textual prompts \cite{ramesh2022hierarchical, saharia2022photorealistic, dong2022dreamartist, nichol2021glide}. A critical challenge known as ``\textit{identity re-contextualization}" (\textit{i.e.}, \textit{ID preservation}, Fig.~\ref{fig:id-fine-edit}, left) \cite{ruiz2022dreambooth, kumari2022multiconcept}, requires modifying a subject's broad context while preserving its identity. However, a more nuanced task, ``\textit{identity fine editing}" (\textit{i.e.}, \textit{ID-context preservation}, Fig.~\ref{fig:id-fine-edit}, right), which demands precise manipulations to a subject's specific features (like lip color or specific accessories), remains underexplored. This task is crucial for applications such as professional photo editing or social networks, where subtle and accurate changes are required without modifying the subject's original identity and context \cite{zhan2021multimodal}.
Despite progress in realistically editing object details \cite{wallace2023edict, huang2023pfb, zhang2023sine}, current models still struggle with detail-oriented human face editing. This difficulty arises due to the complexity of human facial features \cite{deng2018arcface, taigman2014deepface}, the high sensitivity of humans to recognize even minor facial changes \cite{sinha2006face}, and technical challenges in maintaining photorealism during editing \cite{saito2017photorealistic}. These factors make face editing more challenging than editing other objects or scenes.

In ``\textit{identity fine editing}", a key aspect is to preserve the subject's inherent identity and source context. We argue that the source textual prompt is essential to achieve this goal, particularly for reconstructing and safeguarding the conceptual essence of the image. However, current methods, which focus on editing images using target textual prompts, often struggle to preserve both identity and context \cite{mokady2023null, ruiz2022dreambooth, chen2023dreamidentity}. This imbalance between source and target prompts can lead to the loss of parts that do not need to be edited, underscoring the need for a more nuanced approach that better guides the diffusion process for ``\textit{identity fine editing}".
Another key aspect of ``\textit{identity fine editing}" is the ability to make precise manipulations. GAN-based methods propose to disentangle attributes of human faces \cite{harkonen2020ganspace, patashnik2021styleclip, gal2021stylegan}, showing precise editing effects. However, these methods rely on delicate control of hyperparameters or training extra networks to provide guidance. Furthermore, current diffusion-based T2I methods focus on attention mechanisms and rely on manual selections of stages for controllable editing \cite{hertz2022prompt, tumanyan2023plug, cao2023masactrl}. They often overlook the role of noises in the diffusion process, which carries crucial information about varying distributions and interpretation of latent codes \cite{ho2020denoising}. We posit that predicted noises in the sampling process \cite{song2020denoising} are informative and help develop a method to intuitively discern the appropriate stages for precise, disentangled editing.

DreamSalon introduces a staged and noise-guided editing framework that adeptly balances image manipulations and \textit{ID-context preservation}. Initially, DreamSalon discerns aggressive-editing and quality-boosting stages using the frequency and gradient of predicted noises. In the editing stage, it uses this frequency information to guide a weighted mix of source and target text embeddings, primarily focusing on target prompts for detailed editing. Then in the boosting stage, its focus is shifted to the source prompt to maintain the original image's identity and context. As the process moves into the boosting stage, it employs stochastic denoising to enhance the quality of details. Staged editing ensures balance in both specific manipulations and \textit{ID-context preservation}. Additionally, in the editing stage, DreamSalon employs semantic mixing of source and target prompts, leveraging the differences in their embedding covariances for ``identity fine editing." This approach enables precise, context-sensitive adjustments by capturing token variances and relationships, ensuring targeted and nuanced image manipulations.
Experiments demonstrate that DreamSalon can edit specific details on human faces efficiently and faithfully.
Our contributions are as follows:
\begin{enumerate}
    \item \textbf{Challenges}: We identify challenges of an underexplored T2I task, ``\textit{identity fine editing}", which requires manipulations of specific features and \textit{ID-context preservation}.
    \item \textbf{Techniques}: We present a staged, noise-guided editing framework, \textbf{DreamSalon}. It leverages high-frequency information for detailed image manipulations in the editing stage and stochastic denoising for image quality improvement in the boosting stage. Furthermore, its adaptive and semantic mixing of source and target prompts balances the editability and \textit{ID-context preservation}.
    \item \textbf{Superiority}: Experiments show the superiority of DreamSalon in editing facial details, surpassing existing methods both qualitatively and quantitatively.
\end{enumerate}

\section{Related Work}
\label{sec:formatting}

\subsection{Text-to-Image Generation}
Text-conditioned image generation has seen significant advancements recently. Previously, GAN-based \cite{ruan2021daegan, tao2020dfgan, cheng2020rifegan, Li_Qi_Lukasiewicz_Torr_2019} and VAE-based models \cite{esser2021taming, chang2023muse} gain lots of interest at creating high-quality and diverse images. However, these models often struggle to accurately reflect user descriptions and demand substantial optimization time. In comparison, diffusion models stand out for their exceptional semantic understanding and capability to generate varied, photorealistic images directly from textual prompts, offering a distinct advantage in controllability and image quality \cite{yu2022scaling, ramesh2021zero, balaji2022ediffi, kim2022diffusionclip}.

\subsection{Personalized Image Synthesis for Face Identity}

Recent works in personalization have shown significant promise in the realm of generating customized concepts \cite{gal2022image, kumari2022multiconcept, ruiz2022dreambooth, ruiz2023hyperdreambooth}, 
but their extensive optimization requirements limit broader applications. Meanwhile, other works have turned to train extra encoders as a means to efficiently synthesize personalized images \cite{shi2023instantbooth, wei2023elite, ma2023unified, chen2023dreamidentity}. Unlike methods that primarily concentrate on preserving identity while altering context significantly, our work is centered on ``\textit{identity fine editing}", aiming to make precise manipulations on specific features while preserving both identity and context. A concurrent work \cite{wu2023uncovering} utilizes a weighted-mixing strategy to disentangle the editing targets. However, this method fails to capture the semantic relation between prompts during mixing and overlooks the requirements of adaptive editing in the diffusion process. In contrast, our approach distinguishes itself by utilizing the inherent capabilities of diffusion models, rather than training extra encoders. Moreover, it allows for detailed image manipulations without extensive optimization, leading to more efficient personalization.
\begin{figure*}[h]
    \centering
    \includegraphics[width=1.0\linewidth, height=0.33\linewidth]{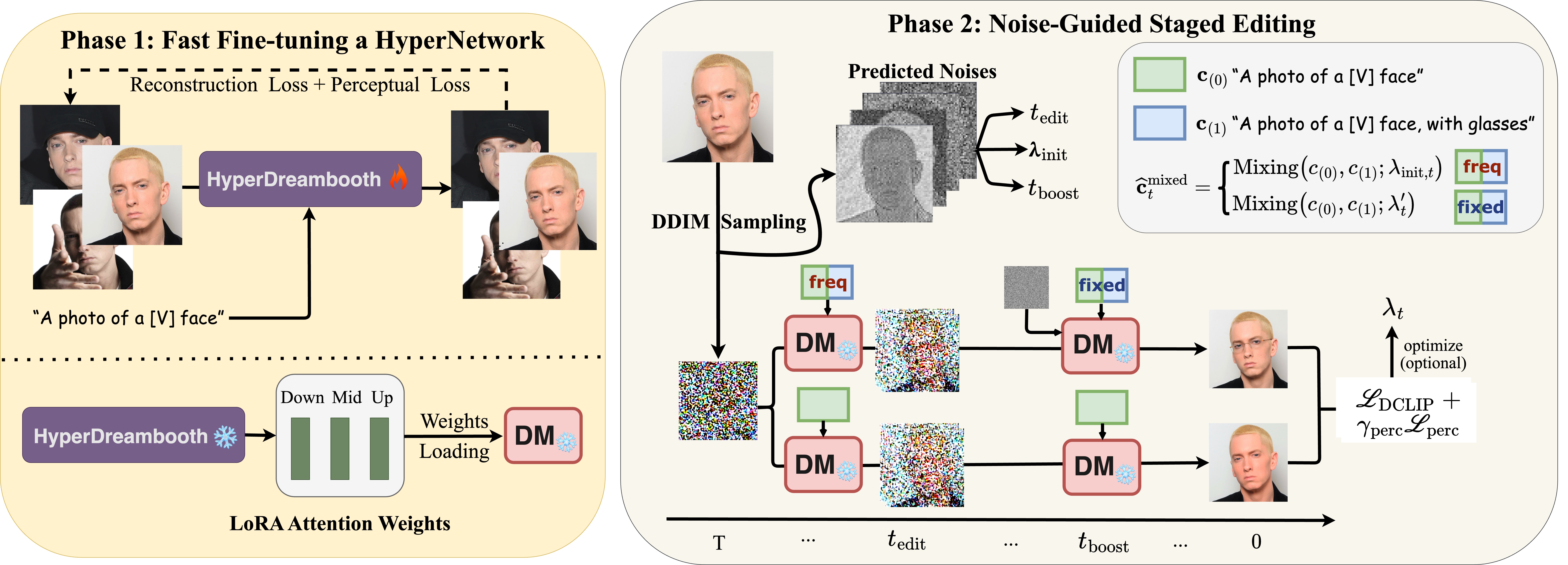}
    \caption{\textbf{DreamSalon pipeline}. Phase 1: fast fine-tuning a hypernetwork per identity, obtaining personalization weights for the Latent Diffusion Model. Phase 2: noise-guided staged editing, where the aggressive-editing stage (before $t_{\text{edit}}$) and the quality-boosting stage (after $t_{\text{boost}}$) are discerned via predicted noises.}
    \label{fig:adaptive-edit}
\end{figure*}

\section{Methods}
\label{sec:methods}

Our framework focuses on fast, fine editing and \textit{ID-context preservation} (Fig.~\ref{fig:adaptive-edit}). It begins by generating personalized weights for each identity. Once these weights are loaded into a pre-trained text-to-image model (Sec.~\ref{subsec:T2I Model}), a noise-guided, staged-editing approach is employed for detailed editing and \textit{ID-context preservation} (Sec.~\ref{subsec:staged-editing}). Additionally, more precise editing is achieved by semantically mixing source and target prompts (Sec.~\ref{subsec:CovGuidance}). 

\subsection{Preliminary} \label{subsec:T2I Model}

\subsubsection{Denoising Diffusion Implicit Model (DDIM)}
DDIM \cite{song2020denoising} redefines DDPM \cite{ho2020denoising} as 
$q_\sigma (\mathbf{x}_{t-1} | \mathbf{x}_t, \mathbf{x}_0)$
a non-Markovian process. DDIM's reverse process reads:
\begin{equation}\label{eq:ddim}
    \begin{split}
        \mathbf{x}_{t-1} &= 
        \sqrt{\alpha_{t-1}}  
        \underbrace{
        \Big(  
            \frac{\mathbf{x}_t - \sqrt{1 - \alpha_t} \epsilon_t^{\theta}(\mathbf{x}_t)} {\sqrt{\alpha_t}}  
        \Big)}_{\text{predicted }\mathbf{x}_0 (\textbf{P}_t(\epsilon_t^\theta(\mathbf{x}_t)))} \\  
        &+ \underbrace{\sqrt{1 - \alpha_{t-1} - \sigma_t^2} \cdot \epsilon_t^{\theta}(\mathbf{x}_t)}_{\text{direction pointing to }\mathbf{x}_t (\textbf{D}_t(\epsilon_t^\theta(\mathbf{x}_t)))}  
        + \underbrace{\sigma_t \epsilon_t}_{\text{random noise}},
    \end{split}
\end{equation}
where $t$ is the timestep, 
$\alpha_t$ and $\sigma_t$ are the variance schedule, $\epsilon^\theta_t$ is the noise predictor, and $\epsilon_t \sim \mathcal{N}(0, \mathbf{I})$ is standard Gaussian noise. The denoising process in Eq.~\ref{eq:ddim} is stochastic when $\eta = 1$, and is deterministic when $\eta = 0$.

\subsubsection{Personalized Weights Generation}
The Latent Diffusion Model (LDM) \cite{rombach2022high} first transforms an input image $\mathbf{x}$ into a lower-resolution latent space $\mathbf{z}$ via a Variational Auto-Encoder (VAE) \cite{kingma2013auto}. Then a text-conditioned diffusion model \cite{radford2021learning} is trained to generate the latent code of the target image from embeddings of text input $\textbf{c}$. In this work, a fast personalized method, HyperDreambooth \cite{ruiz2023hyperdreambooth}, is used to initialize the attention weights of the LDM.  The optimization of these weights is guided by:
\begin{equation}
    \mathcal{L}_{\text{HyperDreambooth}} = \alpha \mathbb{E}_{\epsilon,\mathbf{z},\mathbf{c}} [||\epsilon - \epsilon^\theta (\mathbf{z}_t, \mathbf{c})\|^2_2] + ||\hat{\theta} - \theta ||_2^2,
\end{equation}
where $\hat{\theta}$ are the pre-optimized weight parameters of the personalized model for image $\mathbf{x}$, and $\alpha$ is the hyperparameter that controls the relative weight for two loss terms. This process resembles Dreambooth but is significantly faster (6x) and more storage-efficient (10x). By using 2$\sim$4 images of an identity, we obtain personalized weights (Fig.~\ref{fig:adaptive-edit}, left). These weights are then loaded to the LDM, for the generation of various customized images for that identity (Fig.~\ref{fig:adaptive-edit}, right).

\subsection{Staged Editing}\label{subsec:staged-editing}

In image editing using diffusion models, an adaptive approach to the denoising process is critical, due to the varied distributions of intermediate latent codes \cite{kwon2022diffusion}. We advocate for aggressive edits in the early stage, followed by quality boosting in the later stage. However, identifying the optimal timing for each stage remains under-researched. This section explores how to perform adaptive, staged editing. 

\subsubsection{Boosting Stage with Stochastic Denoising}

To boost the images' quality, we follow the insights of previous work \cite{karras2022elucidating} to employ stochastic denoising instead of the deterministic one (Eq.~\ref{eq:ddim}). However, indiscriminate noise addition throughout the diffusion process may introduce error accumulation, which modifies content significantly, resulting in undesired editing. We posit that the gradient of predicted noise plays an important role in identifying the appropriate timing ($t_{\textbf{boost}}$) to add noise. Since when latents' gradients are small, their value changes become more consistent, indicating a lower risk of major changes to the content. This stage (e.g., 25\% quantile of all gradients, timesteps 20 ($t_\text{boost}$)$\sim$0 in Fig.~\ref{fig:phase-edit}) is ideal for applying Gaussian noise:
\begin{equation*}
    \mathbf{z}_{t-1}=\sqrt{\alpha_{t-1}} \mathbf{P}(\epsilon_t^\theta (\mathbf{z}_t,\mathbf{c}))
    + \mathbf{D}(\epsilon_t^\theta(\mathbf{z}_t,\mathbf{c})) 
    + \mathbf{I}_{[t < t_\text{boost}]}(\sigma_t \epsilon_t),
\end{equation*}
where $\mathbf{I}$ denotes an indicator function for noise addition.

\subsubsection{Editing Stage with Frequency Guidance}

\begin{figure}[t]
    \centering
    \includegraphics[width=1.0\linewidth, height=0.6\linewidth]{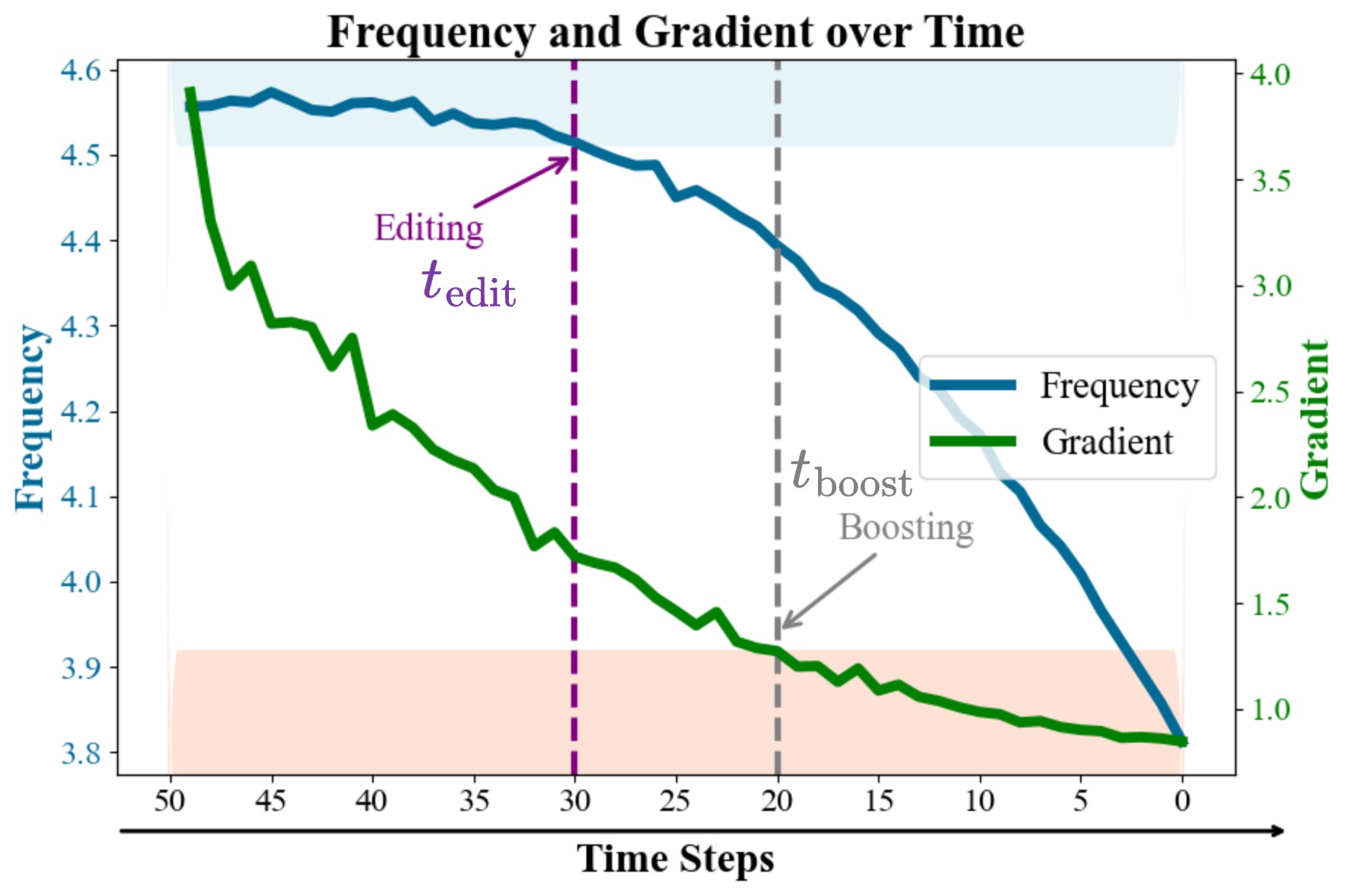}
    \caption{Stage discernment based on frequency and gradient of predicted noises. The editing stage is determined by high-frequency predicted noises (75\% quantile), followed by the boosting stage where gradients are relatively smaller (25\% quantile). More details about the frequency and gradient of predicted noises are available in the \textit{Suppl}.}
    \label{fig:phase-edit}
\end{figure}

DreamSalon discerns the editing stage for manipulations on specific and detailed features, by identifying the high-frequency predicted noises (e.g., 75\% quantile of all frequencies, timesteps 50$\sim$30 ($t_\text{edit}$) in Fig.~\ref{fig:phase-edit}). As higher frequencies correlate strongly with the finer aspects of an image and signify rapid intensity changes \cite{gonzales1987digital}, intermediates with high frequency are ideal for editing without impacting the overall structure, offering efficient ways to alter specific details. Conversely, low-frequency components, indicative of broader, uniform areas, are more influential in modifying the overall appearance rather than specific details. To achieve adaptive control during the diffusion process, we start with the mixing of source and target prompts. Since our goal is to edit the specific details while preserving identity and context, amplifying the contribution of the target prompt during the editing stage aids the manipulation of specific details a lot. This mixing is controlled by a weighting strategy:
\begin{equation}\label{eq:mixing}
    \mathbf{c}^{\text{mixed}}_t = (1-\lambda_t) \mathbf{c}_{(0)} + \lambda_t \mathbf{c}_{(1)},
\end{equation}
where $\mathbf{c}_{(0)}$ and $\mathbf{c}_{(1)}$ represent the CLIP embeddings of the source and target prompts, respectively. The weighting factor $\lambda_t$ at each time step $t$, is the only (optional) parameter optimized in our framework. $\mathbf{c}^{\text{mixed}_t}$ are mixed text embeddings, which balances the impact of the source and target prompts on the generation of edited images. 

For ``\textit{identity fine editing}", our intuition is using the source prompt to help guide the diffusion model to maintain the identity and context, while incorporating the detailed edits suggested by the target prompt. Thus, in the editing stage, DreamSalon puts greater weights on target prompts for aggressive editing (\textit{i.e.}, $\lambda_t$ should be larger). And after that, it focuses on source prompts to retain the core conceptual content of the original image (\textit{i.e.}, $\lambda_t$ should be set as relatively smaller). However, the initialization of the weight factors $\lambda_t$ is intricate. Inappropriate initialization can degrade the editing performance and increase optimization time significantly. Rather than manually setting these factors, we propose to use the inherent attributes of intermediates $\mathbf{x}_t = \psi(\mathbf{z}_t)$ from the DDIM sampling process, where $\psi$ denotes the pretrained decoder. Recall that $\mathbf{x}_t$ with high frequency is mostly present at the early stage, indicating that more aggressive editing is required at these timesteps. Therefore, the initialized weight factors and the mixed text embeddings can be represented as:
\begin{subequations}\label{eq:staged_embds}
    \begin{gather}
        \boldsymbol{\lambda}^{\text{init}} = 
        \begin{cases}
            \text{Normalize}(\text{FFT}(\psi(\mathbf{z}_t)))  & \text{if }  t \le t_{\text{edit}}, \\
            \lambda^\prime & \text{otherwise},
        \end{cases}
        \label{eq:init_weights}
        \\
        \mathbf{c}^{\text{mixed}}_t = (1-\boldsymbol{\lambda}^{\text{init}}_t) \mathbf{c}_{(0)} + \boldsymbol{\lambda}^{\text{init}}_t \mathbf{c}_{(1)},
        \label{eq:adaptive_embds}
    \end{gather}
\end{subequations}
where FFT represents the Fast Fourier Transform, $\boldsymbol{\lambda}_{\text{init}}$ is a vector composed of all $\lambda_t$, and $\lambda^\prime$ is a hyperparameter used after the editing stage. We set $\lambda^\prime$ as 0.2 in our experiments, to leverage the source prompt to improve the reconstruction and preserve identity and context better. 

\subsection{Covariance Guidance for Detailed Editing}  \label{subsec:CovGuidance}

\begin{figure}[t]
    \centering
    \includegraphics[width=1.0\linewidth, height=0.7\linewidth]{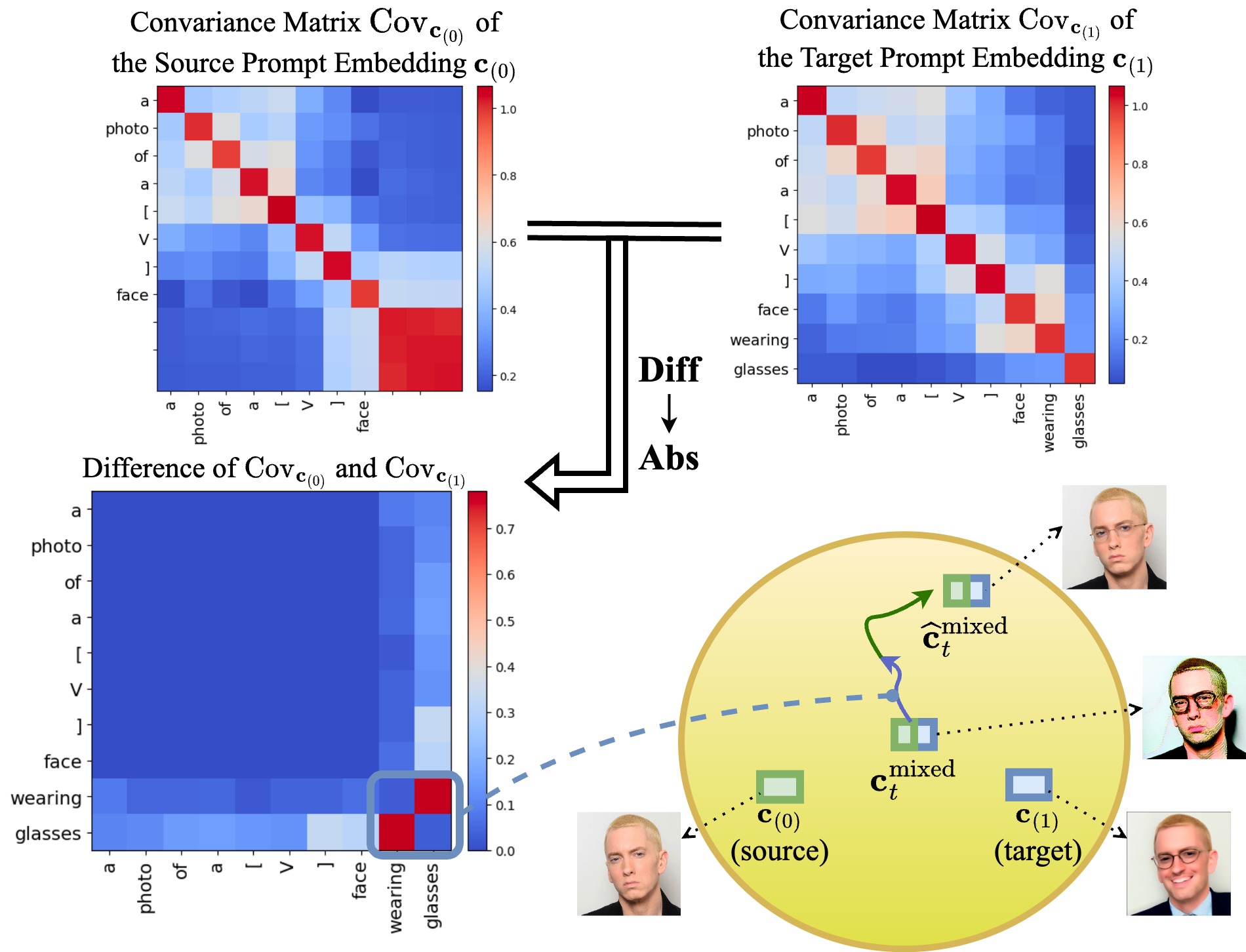}
    \caption{Covariance analysis in prompt embeddings: differences in covariance matrices for source and target prompt embeddings, guide the semantic mixing of prompts for precise attribute editing in generated images.}
    \label{fig:emb_guidance}
\end{figure}

For sensitive tasks like facial attribute editing, directly mixing text embeddings using a weighted sum (Eq.~\ref{eq:adaptive_embds}) can not effectively capture the expressiveness of source and target prompts. This is because a weighted sum uniformly treats each token, failing to account for its specific roles and context. To address this, we propose a covariance guidance method. Since the covariance matrix can assess how variations in one token of the embedding relate to another (Eq.~\ref{eq:cov_mat}), it helps develop a more context-aware integration of prompts \cite{webster1992tokenization}. In particular, by comparing the covariance matrices of source ($\text{Cov}_{\mathbf{c}_{(0)}}$) and target ($\text{Cov}_{\mathbf{c}_{(1)}}$) embeddings, we identify tokens that significantly contribute to desired changes from the difference between $\text{Cov}_{\mathbf{c}_{(0)}}$ and $\text{Cov}_{\mathbf{c}_{(1)}}$, as presented in Fig.~\ref{fig:emb_guidance}. Tokens with significant differences are marked in red, indicating key tokens where the target prompt diverges from the source, thus requiring more attention in the editing process. To utilize this information to guide the diffusion process, we present CovDiff as Eq.~\ref{eq:covdiff}, which acts as a metric for token-specific guidance. It emphasizes tokens in the target prompt that bring desired manipulations, enabling precise editing control. 
\begin{subequations}
    \label{eq:cov}
    \begin{gather}
        \text{Cov}_{\mathbf{c}_{(0)}} = \mathbf{c}_{0} \mathbf{c}_{(0)}^T / (n_{(0)} - 1)
        \label{eq:cov_mat}
        \\
        \text{CovDiff} = \text{Normalize}(\max_{i \text{ or } j}|\text{Cov}_{\mathbf{c}_{(1)}} - \text{Cov}_{\mathbf{c}_{(0)}}|)
        \label{eq:covdiff}
    \end{gather}
\end{subequations}

In Eq.~\ref{eq:cov}, $n_{(0)}$ is the dimension of position tokens and $\max_{i \text{ or } j}$ represents maximization via the x-axis or y-axis since the covariance matrix is symmetric. CovDiff consists of values in the range (0,1), as many as there are tokens.
For more detailed image editing, tokens with higher values in CovDiff should be prioritized in the mixing of text embeddings. We offer token-level, semantic control for improved editability by integrating CovDiff with Eq.~\ref{eq:adaptive_embds} as follows:
\begin{equation}\label{eq:adaptive_embds_with_guidance}
    \mathbf{\hat{c}}_t^{\text{mixed}} =
    \begin{cases}
        \text{CovDiff} \odot \big( (1-\boldsymbol{\lambda}^{\text{init}}_t) \mathbf{c}_{(0)} + \boldsymbol{\lambda}^{\text{init}}_t \mathbf{c}_{(1)} \big) & \text{if }  t \le t_{\text{edit}}, \\
         (1-\boldsymbol{\lambda}^{\text{init}}_t) \mathbf{c}_{(0)} + \boldsymbol{\lambda}^{\text{init}}_t \mathbf{c}_{(1)} & \text{otherwise},
    \end{cases}
\end{equation}
where $\odot$ denotes the Hamilton multiplication. By multiplying CovDiff with the adaptive weighted sum of source and target embeddings, we can modulate the editing intensity of each token to control the contribution of each token in the mixed text embedding. This allows the model to perform more precise manipulations on target features based on semantic guidance.

\subsection{Overall Loss}
Instead of using the initialized weight factors $\lambda_\text{all}$, we provide an option to optimize them by combining directional CLIP loss and perceptual loss. With CLIP's image encoder $\mathcal{E}_I$ and text encoder $\mathcal{E}_T$ \cite{radford2021learning}, directional loss with cosine distance achieves homogeneous editing without mode collapse \cite{gal2021stylegan}:
\begin{equation}
    \mathcal{L}_{\text{DCLIP}} (\mathbf{x}_{\text{edit}}, y_{\text{target}}; \mathbf{x}_\text{source}, y_\text{source}) = 1 - \frac{\Delta \mathbf{I} \cdot \Delta \mathbf{T}} {|| \Delta \mathbf{I}|| ||\Delta \mathbf{T}||},
\end{equation}
where $\Delta \mathbf{T} = \mathcal{E}_T (y_\text{target}) - \mathcal{E}_T (y_\text{source})$ and $\Delta \mathbf{I} = \mathcal{E}_I(\mathbf{x}_\text{edit}) - \mathcal{E}_I (\mathbf{x}_\text{source})$ with source and edited image ($\mathbf{x}_\text{source}$, $\mathbf{x}_\text{edit}$), source and target prompts ($y_\text{source}$, $y_\text{target}$). 

For edits that require identity preservation, we use the perceptual loss \cite{johnson2016perceptual} defined in Eq.~\ref{eq:perc}, to prevent drastic changes in semantic content. Our total loss function is a weighted combination of these losses:
\begin{subequations}
    \label{eq:perc_loss_all_loss}
    \begin{gather}
        \mathcal{L}_{\text{perc}}(\mathbf{x}_{0}^s, \mathbf{x}_{0}^{t} ) = ||\phi(\mathbf{x}_{0}^s ) - \phi(\mathbf{x}_{0}^t) ||_1,
        \label{eq:perc}
        \\
        \mathcal{L}_{\text{total}}(w) = \mathcal{L}_{\text{DCLIP}}(w) + \gamma_{\text{perc}} \mathcal{L}_{\text{perc}}(w),
        \label{eq:overall_loss}
    \end{gather}
\end{subequations}
where  $\phi(\cdot)$ denotes a perceptual network that encodes a given image, and $\gamma_{\text{perc}}$ is a weighting hyperparameter.

\section{Experiments}
\label{sec:experiments}

\begin{figure}[t]
    \centering
    \includegraphics[width=1.0\linewidth]{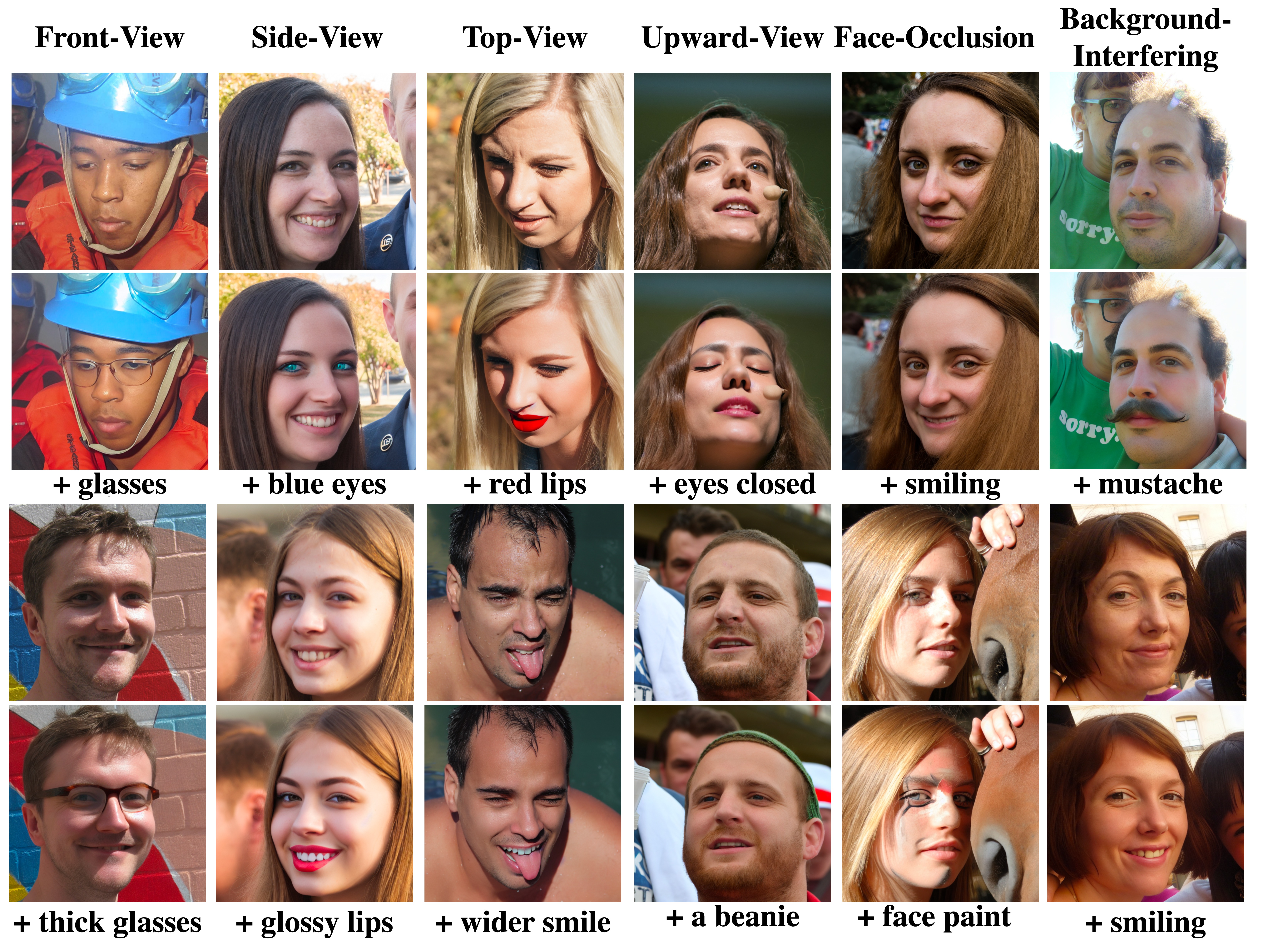}
    \caption{The FFE-Bench for fine-grained face editing across different views and challenging conditions, with  DreamSalon's edits like attribute additions and expression changes.}
    \label{fig:ffe-bench}
\end{figure}

\begin{figure*}[t]
    \centering
    \includegraphics[width=1.0\linewidth]{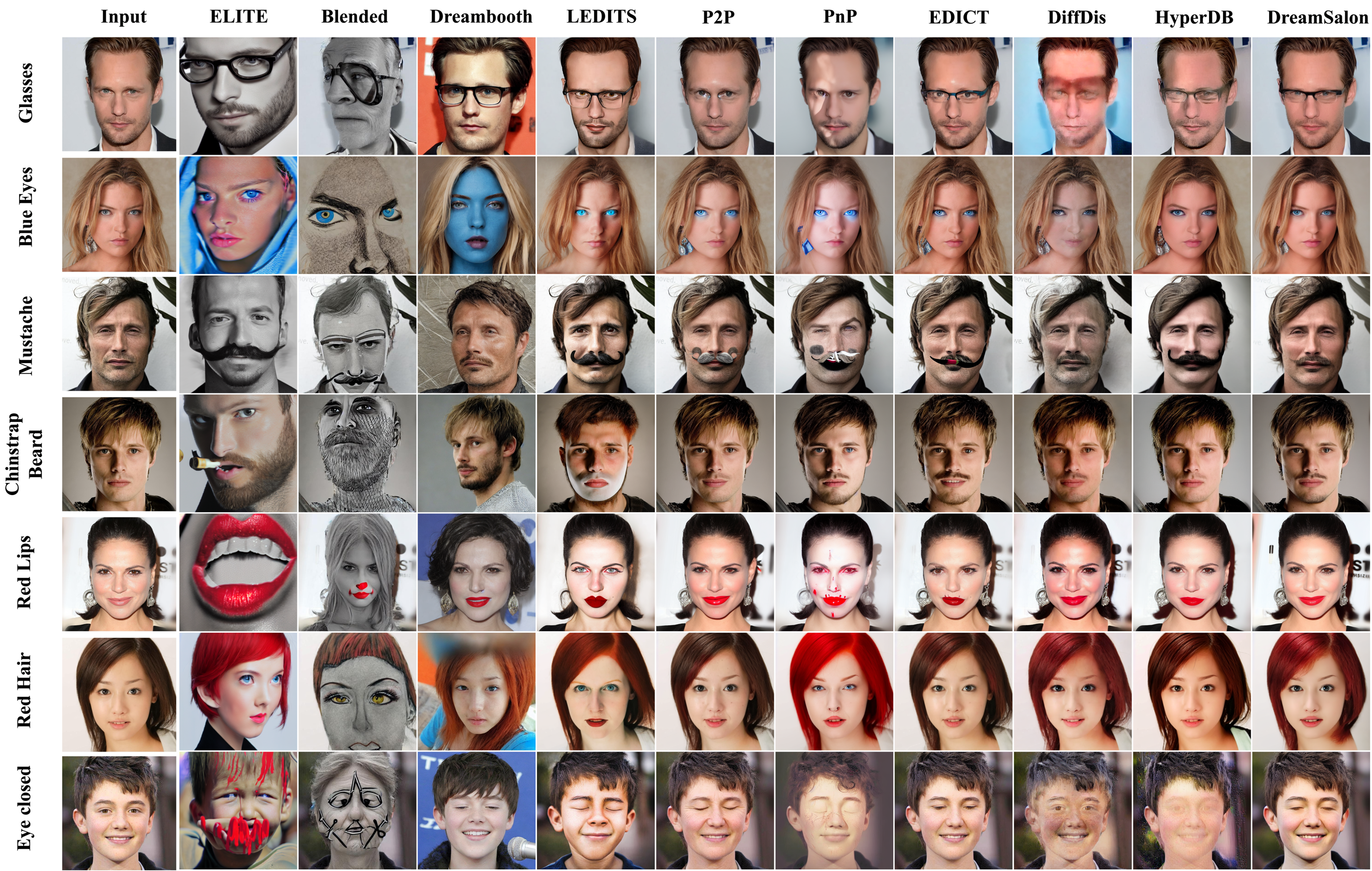}
    \caption{Comparative analysis showcases DreamSalon's precision in aligning images with text descriptions while maintaining identity and context, surpassing other leading methods.}
    \label{fig:comp_overall}
\end{figure*}

\subsection{Experimental Settings}

\textbf{Datasets.} \ We first conduct experiments on the CelebA-HQ dataset \cite{karras2017progressive}. Following existing works \cite{ruiz2022dreambooth, kumari2022multiconcept, wei2023elite}, we use 35 identities for image editing. Each identity is edited with 20 different prompts, and we randomly generated five edited images per identity-prompt combination, totaling 3,500 images. Furthermore, acknowledging the performance on unaligned face images, we construct a \textbf{F}ace-oriented \textbf{F}ine \textbf{E}diting \textbf{Bench}mark dataset from FFHQ-unaligned (\textit{FFE-Bench}, Fig.~\ref{fig:ffe-bench}), which features 600 face images with six types: four views (front, side, top, upward) and two conditions (face-occlusion, background-interfering). Each image is annotated with 20 source/target prompts, editing instructions, and an editing mask for metric calculations.

\noindent\textbf{Evaluation Metrics.} \ Following Dreambooth \cite{ruiz2022dreambooth}, we evaluate our method with three metrics: \textit{CLIP-I}, \textit{CLIP-T}, and \textit{DINO-I}, which assess visual similarity, text-image alignment, and identity uniqueness, respectively. For \textit{CLIP-I}, we calculate the CLIP visual similarity between the source and the generated images. For \textit{CLIP-T}, we calculate the CLIP text-image similarity between the generated images and the text prompts given. For \textit{DINO-I}, we calculate cosine similarity between the ViT-B/16 DINO \cite{caron2021emerging} embeddings of source and generated images. Moreover, we adopt fine-tuning and editing time as a metric to evaluate the efficiency.

\noindent\textbf{Implementation Details.} \ We choose SD 1.5 as our base T2I model. During obtaining the weights using HyperDreambooth \cite{ruiz2023hyperdreambooth}, the learning rate is set to 5e-5, the embedding regularization weight is set to 1e-4. The weights are generated with 2$\sim$4 images for each identity. During inference, we use the PLMs sampler \cite{liu2022pseudo} with 50 timesteps, and the scale of classifier-free guidance is 5. For adaptive editing, we use $\lambda^\prime$ = 0.2 after the editing stage. Optional optimization is performed for 3 iterations with a learning rate of 5e-2.

\subsection{Experimental Results}

We present both qualitative and quantitative results to highlight our method's superiority in ``\textit{identity fine editing}" over current SOTA diffusion-based models. Our comparisons include both fine-tuning methods \cite{ruiz2022dreambooth, ruiz2023hyperdreambooth} and fine-tuning free methods \cite{hertz2022prompt, tsaban2023ledits, wallace2023edict, wei2023elite, tumanyan2023plug, avrahami2023blended, wu2023uncovering}, in which some methods require additional guidance, such as masks \cite{wei2023elite, avrahami2023blended}. 

\subsubsection{Qualitative Results}

Our method, DreamSalon, precisely edits image details according to text prompts, demonstrating versatility in various edits like adding accessories, altering facial hair, or changing eye color, as shown in Fig.~\ref{fig:ffe-bench},~\ref{fig:comp_overall}. A critical aspect of our method's performance is its ability to preserve both identity and context in the images, avoiding altering prompt-unrelated image aspects, and ensuring detail-specific changes. In comparison, ELITE and Blended Latent Diffusion often change both identity and context, Dreambooth alters context, and LEDITS maintains context but changes identity. Other methods either introduce undesired changes or fail to make prompt-specific edits. For instance, PnP and P2P inaccurately edit earring colors when prompted for eye color changes, showing less control (2nd row); HyperDreambooth and EDICTS struggle with hair color changes (6th row). DreamSalon's edits maintain realism, blending seamlessly with original image characteristics like lighting and texture, in contrast to the stylistic, lighting, and texture changes seen in Blended Latent Diffusion and LEDITS. More qualitative results are in the \textit{Suppl}.

\subsubsection{Quantitative Results}

As presented in Tab.~\ref{table: performance_comp}, DreamSalon excels in our performance comparison, achieving the highest CLIP-I score, indicating superior \textit{ID-context preservation} on face images during editing. This means that both identity and context in edited images closely match their originals. DreamSalon also leads in CLIP-T scores (0.247), reflecting its accuracy in mirroring text prompts in image edits, demonstrating a robust understanding and application of textual instructions. Additionally, it tops in DINO-I scores, maintaining conceptual similarity with target images. Regarding time efficiency (Tab.~\ref{table: combined_time_comparison}), DreamSalon, assisted by HyperDreambooth, fine-tunes 6x faster and requires 10x smaller storage than Dreambooth. After fine-tuning, DreamSalon's editing time is slightly longer than Dreambooth and HyperDreambooth, but its editing performance significantly outdoes them. Among fine-tuning-free methods, ELITE is the fastest, while DreamSalon matches this editing speed. Other methods like P2P, PnP, and LEDITS take longer to edit than DreamSalon. When making multiple different edits per identity, the time cost of fine-tuning is negligible.

\begin{table}[t]
\centering
\caption{Quantitative comparisons on CelebA-HQ and our FFE-Bench. Our method outperforms SOTA T2I methods in terms of face similarity, text-alignment, and conceptual similarity.}
\label{table: performance_comp}
\begin{tabular}{c|c|c|c}
     \hline
     \textbf{Method} & \textbf{CLIP-I} $\uparrow$ & \textbf{CLIP-T} $\uparrow$ & \textbf{DINO-I} $\uparrow$ \\
      \hline
    \multicolumn{4}{c}{\textbf{CelebA-HQ}} \\
    \hline
     DB \cite{ruiz2022dreambooth}     & 0.705 & 0.210 & 0.150 \\
     LEDITS \cite{tsaban2023ledits}   & 0.606 & 0.242 & 0.940 \\
     P2P \cite{hertz2022prompt}   & 0.731 & 0.229 & 0.937 \\
     PnP  \cite{tumanyan2023plug}   & 0.621 & 0.245 & 0.928 \\
     EDICT  \cite{wallace2023edict}  & 0.812 & 0.212 & 0.948 \\
     DiffDis  \cite{wu2023uncovering}  & 0.779 & 0.205 & 0.881 \\
     HyperDB \cite{ruiz2023hyperdreambooth}  & 0.675 & 0.224 & 0.620\\
     \textbf{DreamSalon}     & \textbf{0.837} & \textbf{0.247} & \textbf{0.958} \\
     \hline
     \multicolumn{4}{c}{\textbf{FFE-Bench}} \\
     \hline
     PnP     & 0.603 & 0.238 & 0.924 \\
     EDICT   & 0.808 & 0.204 & 0.898  \\
     HyperDB  & 0.639 & 0.215 & 0.598 \\
     \textbf{DreamSalon}     & \textbf{0.815} & \textbf{0.242} & \textbf{0.932} \\
     \hline
\end{tabular}
\end{table}

\subsection{Ablation Studies}

\begin{table}[t]
\centering
\caption{Ablation study on personalized weight generation, staged editing and covariance-guided prompts mixing.}
\label{table: ablation}
\begin{tabular}{c|c|c|c}
     \hline
    \textbf{Method} & \textbf{CLIP-I} $\uparrow$ & \textbf{CLIP-T} $\uparrow$ & \textbf{DINO-I} $\uparrow$ \\
      \hline
     w/o HyperDB     & 0.778 & 0.224 & 0.847 \\
     w/o Staged   & 0.734 & 0.228 & 0.792\\
     w/o Frequency    & 0.750 & 0.233 & 0.835 \\
     w/o Boosting  & 0.804 & 0.241 & 0.904 \\
     w/o CovDiff    & 0.802 & 0.234 & 0.877 \\
     w/o Opt $\lambda_t$ & 0.812 & 0.239 & 0.939 \\
     \textbf{DreamSalon}   & \textbf{0.837} & \textbf{0.247} & \textbf{0.958} \\
     \hline
\end{tabular}
\end{table}

\begin{figure}[t]
    \centering
    \includegraphics[width=1.0\linewidth]{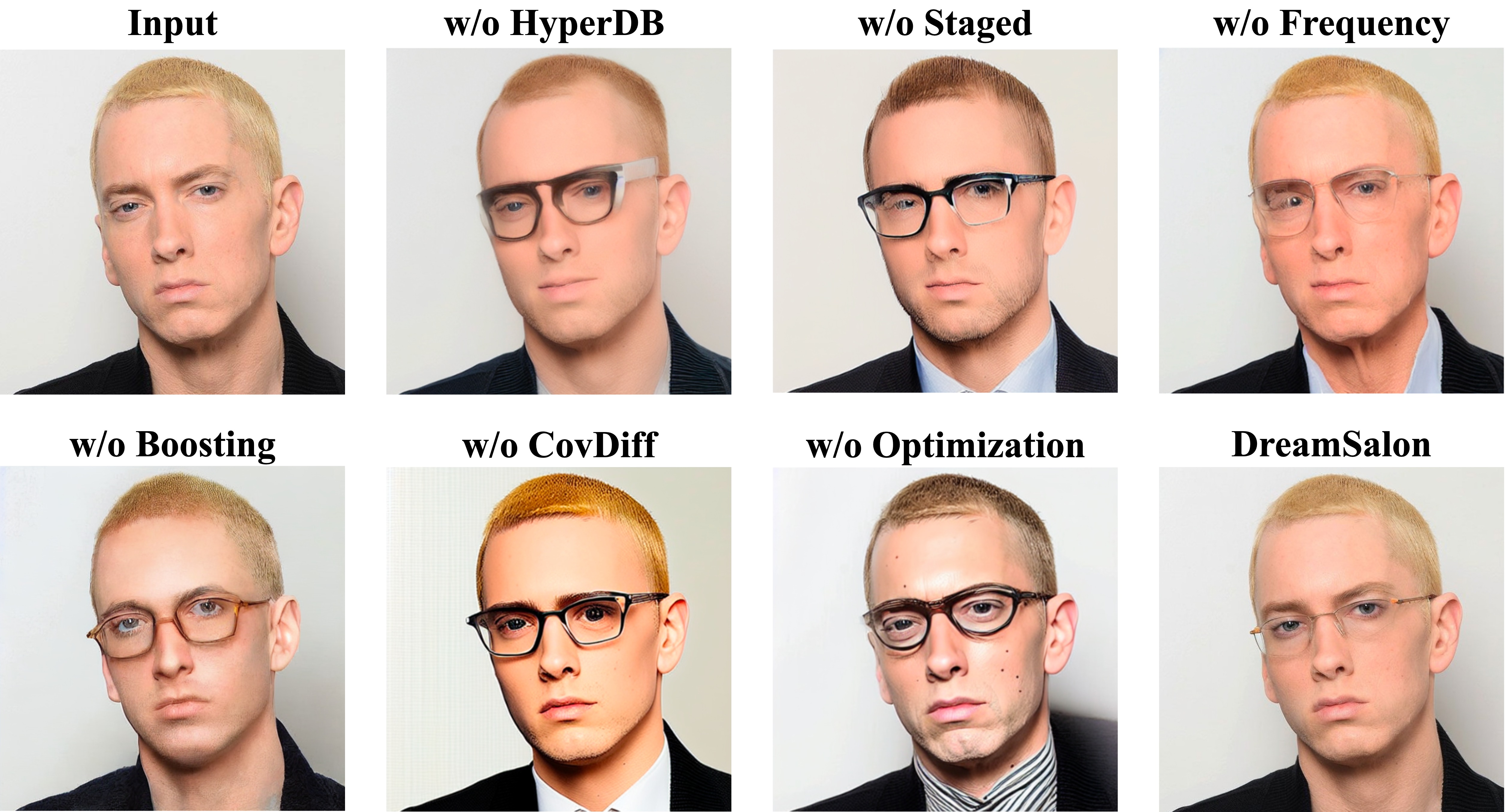}
    \caption{Qualitative comparisons with various components omitted, culminating in the full DreamSalon method which integrates all features for optimal editing outcomes.}
    \label{fig:ablations}
\end{figure}

Ablation studies, as seen in Fig.~\ref{fig:ablations} and Tab.~\ref{table: ablation}, demonstrate the impact of each method on the performance of face editing. ``w/o HyperDB" indicates without using HyperDreambooth to fine-tune, showing reduced identity preservation. ``w/o Frequency" indicates using a prefixed value for $\lambda_t$ of all timesteps, which leads to a loss of specific details like mouth and collar, indicating the need for a dynamic value for $\lambda_t$ at different timesteps. ``w/o Boosting" denotes the use of deterministic denoising instead of stochastic one during the boosting stage, resulting in the loss of finer details such as eyes. ``w/o Staged" condition is the combination of ``w/o Frequency" and ``w/o Boosting". ``w/o CovDiff" alters overall texture and saturation due to the absence of semantic, detailed guidance in mixing text embeddings. Finally, ``w/o Opt $\lambda_t$" introduces some artifacts. DreamSalon outperforms all ablated versions in all metrics, affirming the efficacy of its entire method.

\begin{table*}[t]
\centering
\caption{Comparison of fine-tuning time and editing time with fine-tuning free and fine-tuning based methods.}
\begin{tabular}{l||c c c c c c|c c c}
\hline
{\multirow{2}{*}{Methods}} & \multicolumn{6}{c|}{Fine-tuning Free Methods}  & \multicolumn{3}{c}{Fine-tuning based Methods} \\
\cline{2-10}            & Blended & P2P & PnP & LEDITS & EDICT & ELITE & Dreambooth & HyperDB & DreamSalon \\ \hline
Editing Time (s)        & 62       & 97  & 65  & 48     & 42    & 28    & 14         & 14         & 26         \\ \hline
Fine-tuning Time (s)    & \multicolumn{6}{c|}{N/A}                        & 640        & 106         & 106        \\ \hline
\end{tabular}
\label{table: combined_time_comparison}
\end{table*}

\subsection{Experimental Analysis}

\subsubsection{Different Covariance Guidance}
\begin{figure}[t]
    \centering
    \includegraphics[width=1.0\linewidth, height=0.55\linewidth]{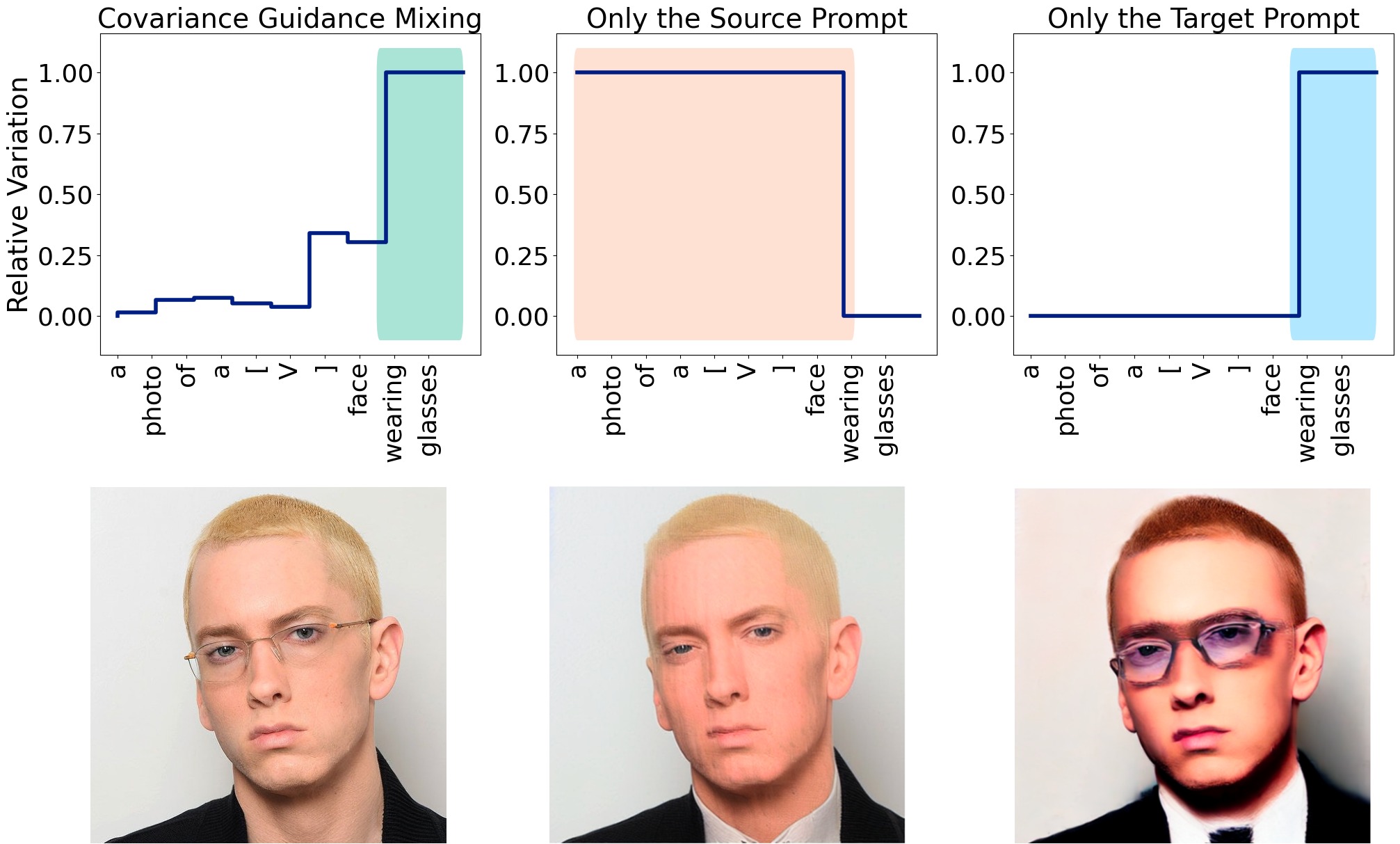}
    \caption{Impact of covariance guidance on prompt mixing: the effect of mixing source and target prompts with covariance guidance versus using only source or only target prompt, and their respective influence on the edited image outcome.}
    \label{fig:covdiff_analysis}
\end{figure}

We provide further analysis of the role of covariance guidance (Eq.~\ref{eq:covdiff}, \ref{eq:adaptive_embds_with_guidance}) in Fig.~\ref{fig:covdiff_analysis}. As stated before, tokens with high values in CovDiff signal its importance for precise edits, and low CovDiff indicates less impact, necessitating balance to avoid overemphasis and maintain overall context. The first column represents the CovDiff used in our methods, considering the difference between the covariance matrices of the source and target embeddings. Tokens that present a larger variance in the target text prompt, as indicated by a larger CovDiff, guide the model to focus its editing efforts, introducing new details (adding glasses), while preserving identity-context. The second column employs an extreme condition where the tokens unique in the target prompt are ignored, the editing fails to correspond to the text instruction in the target prompt. The third column considers another extreme condition, where tokens in the source prompt are ignored, even though editing corresponding to the text instruction in the target prompt is performed, there is a failure to preserve identities and context. 
By focusing on tokens that bring notable changes in the editing stage, DreamSalon ensures precise editing on minor details, enabling the creation of images that are not only visually appealing but also contextually consistent with both source and target prompts.

\subsubsection{Mixed Text Embeddings with Image Embeddings}

To better understand how the adaptive mixing of prompt embeddings contributes to the precise editing, we assess the distance between mixed text embeddings and source/target image embeddings during and after the editing stage, as depicted in Fig.~\ref{eq:staged_embds}. During the editing stage (left), the mixed prompt embeddings closely align with the target image embeddings, indicating heightened editability for detailed modifications. On the contrary, after the editing stage (right), the mixed prompts exhibit closer proximity to the source image embeddings, signifying a greater focus on reconstructing aspects from the source image. This emphasis on source image reconstruction contributes to identity and context preservation after the editing stage.

\begin{figure}[t]
    \centering
    \includegraphics[width=1.0\linewidth, height=0.5\linewidth]{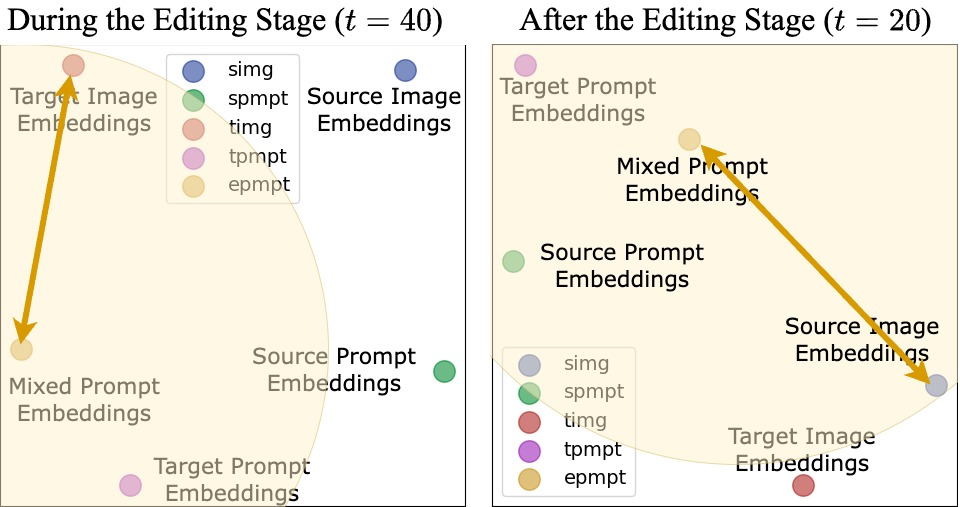}
    \caption{The embeddings of images and prompts during and after the editing stage, highlighting the shift from a focus on target image embeddings to integration with source image embeddings for precise and context-aware image manipulation.}
    \label{fig:stage_embds}
\end{figure}

\subsubsection{Discernment of Different Stages}
\begin{figure}[t]
    \centering
    \includegraphics[width=1.0\linewidth]{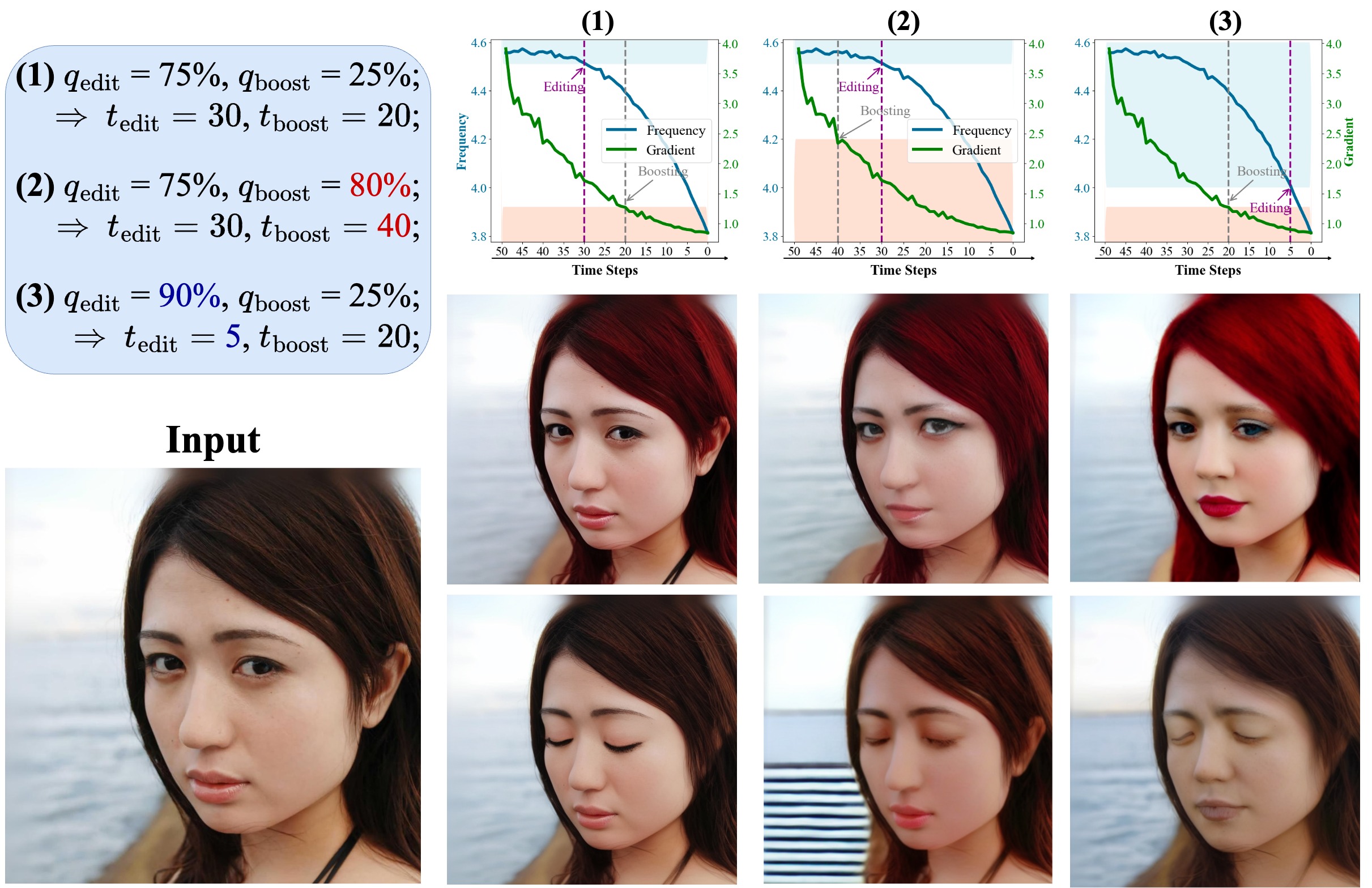}
    \caption{Selections of editing and boosting stages with different quantiles, influencing editing intensities and noise-boosting percentages. The first row depicts varying stage durations, while the second and third rows demonstrate the application of these stages in altering hair color and eye state, respectively.}
    \label{fig:discernment_stages}
\end{figure}

As shown in Fig.~\ref{fig:discernment_stages}, our investigation into discerning the editing and boosting stages involves selecting different quantiles for these phases. The first column adheres to our method's quantile settings, resulting in precise editing while preserving both identity and context. The second column demonstrates a broader range for quality boosting, leading to significant identity or context changes due to excessive noise addition. The third column presents a range for more aggressive editing, where ID-preservation is compromised as a result of insufficient integration with the source prompt, which is essential for maintaining the original essence.

\section{Conclusions}
\label{sec:conclusions}

In summary, DreamSalon offers a framework in ``\textit{identity fine editing}" for text-to-image models, adeptly manipulating specific features while safeguarding the subject's identity and context. It outperforms recent work with a noise-guided, staged-editing framework that precisely manipulates image details through adaptive editing and semantic prompt mixing. Our experiments showcase DreamSalon's exceptional performance in precise and efficient human face editing, marking its advance over existing approaches.

\subsubsection{Ethics Statement}
The adaptability of our approach offers a wide range of opportunities for its use in different areas of structured image creation. Nonetheless, it's crucial to consider the potential societal effects stemming from improper use. Issues such as data leakage could pose privacy risks, and the ability of the model to generate extremely lifelike images might be used for creating content that is illegal or damaging. Adhering to explicit usage policies and committing to its responsible and ethical application is paramount.

\subsubsection{Acknowledgement} 
This work was supported by National Science and Technology Major Project (2022ZD0117102), National Natural Science Foundation of China (62293551, 62377038,62177038,62277042). Project of China Knowledge Centre for Engineering Science and Technology, Project of Chinese academy of engineering ``The Online and Offline Mixed Educational Service System for 'The Belt and Road' Training in MOOC China". ``LENOVO-XJTU" Intelligent Industry Joint Laboratory Project.

\renewcommand\refname{\Large\centerline{References}\global\def\refname{References}}
\bibliographystyle{abbrv}
\bibliography{refs}

\end{document}